\documentclass[conference]{IEEEtran}
\IEEEoverridecommandlockouts
\usepackage{cite}
\usepackage{amsmath,amssymb,amsfonts}
\usepackage{algorithmic}
\usepackage{graphicx}
\usepackage{textcomp}
\usepackage{xcolor}
\usepackage{hyperref}
\usepackage{algorithm}
\usepackage{algorithmic}
\usepackage{booktabs}
\usepackage{multirow}
\usepackage{makecell}
\usepackage{subfigure}
\usepackage{graphicx} 

\def\BibTeX{{\rm B\kern-.05em{\sc i\kern-.025em b}\kern-.08em
    T\kern-.1667em\lower.7ex\hbox{E}\kern-.125emX}}
\begin{document}
\title{FLP-XR: Future Location Prediction on Extreme Scale Maritime Data in Real-time}

\author{\IEEEauthorblockN{George S. Theodoropoulos}
\IEEEauthorblockA{\textit{Dept. of Informatics} \\
\textit{University of Piraeus}\\
Piraeus, Greece \\
gstheo@unipi.gr}
\and
\IEEEauthorblockN{Andreas Patakis}
\IEEEauthorblockA{\textit{Dept. of Informatics} \\
\textit{University of Piraeus}\\
Piraeus, Greece \\
patakisa@unipi.gr}
\and
\IEEEauthorblockN{Andreas Tritsarolis}
\IEEEauthorblockA{\textit{Dept. of Informatics} \\
\textit{University of Piraeus}\\
Piraeus, Greece \\
andrewt@unipi.gr}
\and
\IEEEauthorblockN{Yannis Theodoridis}
\IEEEauthorblockA{\textit{Dept. of Informatics} \\
\textit{University of Piraeus}\\
Piraeus, Greece \\
ytheod@unipi.gr}
}

\maketitle

\begin{abstract}
Movements of maritime vessels are inherently complex and challenging to model due to the dynamic and often unpredictable nature of maritime operations. Even within structured maritime environments, such as shipping lanes and port approaches, where vessels adhere to navigational rules and predefined sea routes, uncovering underlying patterns is far from trivial. The necessity for accurate modeling of the mobility of maritime vessels arises from the numerous applications it serves, including risk assessment for collision avoidance, optimization of shipping routes, and efficient port management. This paper introduces FLP-XR, a model that leverages maritime mobility data to construct a robust framework that offers precise predictions while ensuring extremely fast training and inference capabilities. We demonstrate the efficiency of our approach through an extensive experimental study using three real-world AIS datasets. According to the experimental results, FLP-XR outperforms the current state-of-the-art in many cases, whereas it performs 2-3 orders of magnitude faster in terms of training and inference. 
\end{abstract}

\begin{IEEEkeywords}
Future Location Prediction, Gradient Boosting, Mobility Data Analytics, Maritime Data, Edge Computing
\end{IEEEkeywords}

\section{Introduction}\label{sec:I}

In recent years, the number of smart devices, such as smartphones, wearable devices, has increased significantly\cite{connected_devices}. The widespread presence and convenience of these GPS-enabled devices have led to a growing reliance on them for decision-making, thereby making human life increasingly data-driven. A relevant paradigm is observed in the transport and logistic industries, where maritime vessels, airplanes, and trucks are continuously monitored. The data generated from these monitoring activities are utilized to assess current situations and play a critical role in both short-term and long-term strategic planning for companies. 

As the volume of data generated per second has grown exponentially, the architectures developed to manage this data have also evolved to meet the real-time demands of modern applications. In this context, edge computing\cite{edge_chal} has emerged as a new paradigm, supplementing or even replacing cloud computing in scenarios where data sources are distributed across multiple locations, and low latency is essential. Traditional cloud-based data processing often falls short of meeting these requirements. However, a significant limitation of edge computing is that edge nodes possess more limited storage and processing capacities compared to centralized cloud systems, potentially constraining performance in highly data-intensive applications.

\begin{figure}
    \centering
    \includegraphics[width=0.3\textwidth]{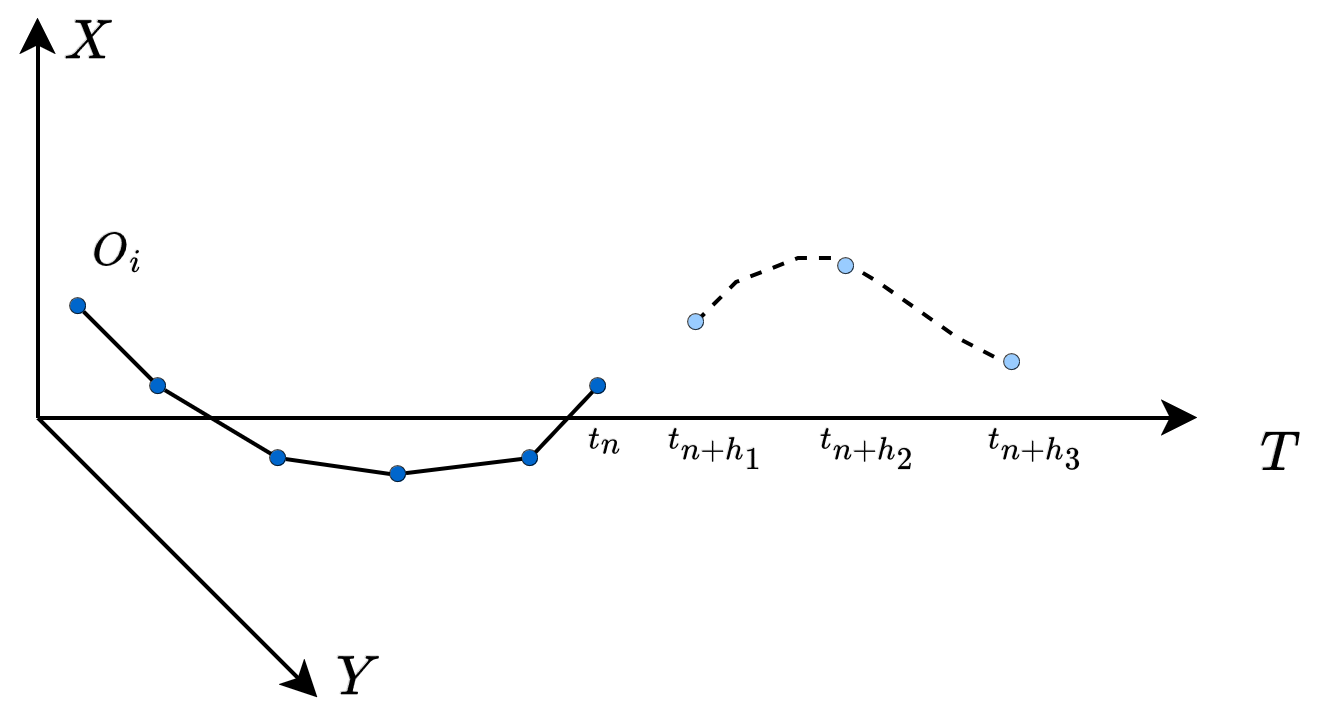}
    \caption{Predicting the future location of vessel $O_i$ for three distinct horizon values after $t_n$.}
    \label{fig:flp_example}
    \vspace{-15pt}
\end{figure}

Predicting the evolution in time of a vessel's trajectory\cite{zhang22} is a critical task in the field of maritime data analytics\cite{artikisbook}, as it serves as a foundational element in various applications, including maritime route modeling\cite{zissis2020}, future collision risk assessment\cite{tritsarolis_collision} and awareness\cite{michalksi}, anomaly detection\cite{pallotta_secuirity}, and many more. In essence, Future Location Prediction (FLP) aims to forecast the anticipated future positions of a moving object (a vessel for the scope of this paper) by analyzing both its individual movement history and broader patterns observed in the movement data of similar objects\cite{flp_review}. In essence, FLP seeks to determine the object's coordinate position at the specified future time horizon. A visual example of an FLP task is illustrated in Figure \ref{fig:flp_example}.

Since the FLP problem can be framed as a time series regression prediction task, approaches including Auto-Regressive Integrated Moving Average (ARIMA) and Markov-based models were originally proposed as candidate solutions\cite{liu}. However, as the authors in\cite{Eva} point out, these conventional approaches struggle to handle the non-linear characteristics of mobility data and the complex, long-term spatial and temporal dependencies inherent to such datasets. 

Nowadays, the majority of approaches in the literature have leveraged various Recurrent Neural Networks (RNN) variants to address the task. By applying these networks in combination with carefully designed preprocessing strategies, state-of-the-art performance can be achieved\cite{tritsarolis2024}\cite{Eva}. However, these models require extensive training on large datasets and are highly computationally intensive, resulting in slow processing speeds during both training and inference, at least in more resource-constrained environments. Specifically, Rezk et al. \cite{rezk} examines the deployment of RNNs on embedded systems and addresses the challenges of optimising these networks within the constraints, and discusses the trade-offs between performance and flexibility. The results of this study highlight the necessity to create more efficient methods tailored for low-resource settings.

Given this context, the requirements for a FLP method in today’s demanding environment are straightforward yet highly challenging: it must be both fast and accurate. Although much of the research emphasizes enhancing the latter, the critical importance of speed is often overlooked. A future-proof FLP method would prioritize both training and inference speed, ensuring rapid processing throughout its lifecycle. It would maintain high-quality performance while being lightweight and adaptable, seamlessly integrating into resource-constrained edge computing units and diverse devices.

To address these challenges and adapt to the new era of large-scale mobility analytics, this paper proposes FLP-XR, a highly efficient \underline{FLP} solution based on the eXtreme Gradient Boosting (XGBoost) model\cite{xgboost}, able to process e\underline{X}treme scale maritime (AIS) data in \underline{R}eal-time . We demonstrate that, in many cases, it is (the lightweight) FLP-XR that outperforms the (resource-demanding) state-of-the-art, while at the same time achieving exceptional efficiency in both training and inference times. Additionally, due to its low training resource requirements, the proposed model can adapt to changes in data efficiently, making it robust even as use-case characteristics evolve.  In essence, we propose a forward-thinking approach to mobility prediction, one that is not only grounded in current needs but also anticipates future developments in the edge / fog / cloud Computing Continuum (CC) field. 

In summary, the main contributions of this work are as follows:
\begin{itemize}
    \item We propose a novel lightweight XGBoost-based approach for future location prediction in the maritime domain, specifically designed for resource-constrained computing environments.
    \item Our method achieves training speeds that are on average 300x faster and inference speeds that are 3000x faster than the state of the art, while maintaining competitive prediction accuracy, outperforming a state-of-the-art model in many cases.
    \item The proposed model's efficiency and compact design enable deployment on edge devices, making it a practical solution for real-time maritime applications where computational resources are limited.
\end{itemize}

The rest of this paper is organized as follows: Section \ref{sec:II} discusses related work. Section \ref{sec:III} formulates the problem at hand and presents our approach for efficient resource-constrained FLP. Section \ref{sec:IV} presents our experimental study over three real-world datasets (namely, Brest, Piraeus, and Aegean) and different hardware settings (from a powerful GPU cluster to a limited resource edge device), assessing the quality and efficiency of our approach. Finally, Section \ref{sec:V} concludes the paper, also providing suggestions for future work.

\section{Related Work}\label{sec:II}

As already mentioned, the earlier approaches in the field tried to handle the FLP problem as a stochastic process, with  Markov models employed to address the prediction challenge. For instance, Liu et al.\cite{liu} introduce L-VTP for predicting the long-term trajectories of ocean vessels. More specifically, they separate a given sea area into a grid and the algorithm leverages multiple sailing-related parameters and a K-order multivariate Markov Chain to build state-transition matrices for trajectory prediction. They use conditional entropy to evaluate their model and perform grid search to identify the best K-order and the most useful sailing parameters.

In a more recent work, Zygouras et al.\cite{zygouras_24} introduce EnvClus$\ast$, an enhanced version of their previous research \cite{zygouras_21}, where a data-driven framework is developed for vessel trajectory forecasting. This method defines the space in which vessels navigate by creating envelopes and corridors, which represent the potential areas of movement. Forecasting is then performed using mobility graphs that model the vessels' most probable future movements, offering a more accurate and structured approach to trajectory prediction.

Another common approach for addressing FLP is by using RNNs like Long-Short Term Memory (LSTM) and Gated Reccurent Unit (GRU) networks that are tailored to sequential data classification and regression.
Nguyen et al.\cite{nguyen} propose a method to predict vessels' destination ports and estimated arrival times using a sequence-to-sequence architecture with LSTM backbone that employs a grid, where each cell is coded to record the sequence of moves. The authors perform beam-search to find not only the most probable trajectory but also alternatives, ranked by their perplexities. The model predicts future vessel movements using a distributed architecture and load balancing for high performance and scalability.
Wang et al.\cite{wang} propose an LSTM-based framework to predict user trajectories, particularly focusing on multi-user and multi-step prediction scenarios relevant to 5G networks. To address limitations in user-specific models, they introduce a region-oriented, Seq2Seq learning approach to improve generalization and reduce error accumulation.
Li et al. \cite{yuhao_li24}, explore the use of a novel hybrid methodology, combining a Graph Attention network (GAT) and an LSTM. The proposed GAT-LSTM model aims to improve the accuracy and robustness of trajectory prediction by considering both spatial and temporal features.

Chondrodima et al. \cite{Eva} present a framework for predicting the locations of vessels up to 60 minutes ahead, even for those not previously recorded. This framework is based on LSTM to address challenges in maritime tracking, such as variable sampling rates, sparse trajectories, and noisy GPS data. A key feature of the framework is its trajectory data augmentation method, which enhances predictive accuracy.
Extending this work, Tritsarolis et al. \cite{tritsarolis2024} propose a new system for predicting the location of vessels based on AIS data. The system, called Nautilus, utilizes LSTM neural networks and outperforms existing methods for short-term prediction. The paper also explores the application of federated learning (FL) to the system, resulting in a new architecture called FedNautilus. In this paper, we consider the (centralized) Nautilus model as the state-of-the-art (SotA) model, which we compare our proposal with.

In resource-constrained environments. Liang et al.\cite{liang} presents a novel GPU-accelerated method, for compressing and visualizing large-scale vessel trajectories derived from AIS data within Maritime Internet of Things (IoT) systems. The algorithm enhances traditional Kernel Density Estimation (KDE) and by incorporating GPU acceleration significantly speeds up processing, enabling real-time visualization. Experiments using multiple datasets demonstrate the superior performance and efficiency of the GPU-implementations compared to traditional KDE-based methods\cite{kdes}. To the best of our knowledge, there is no existing work that specifically addresses the FLP problem within IoT and resource-constrained environments in the maritime domain.

\begin{figure*}[t]
    \centering
    \includegraphics[width=0.8\textwidth]{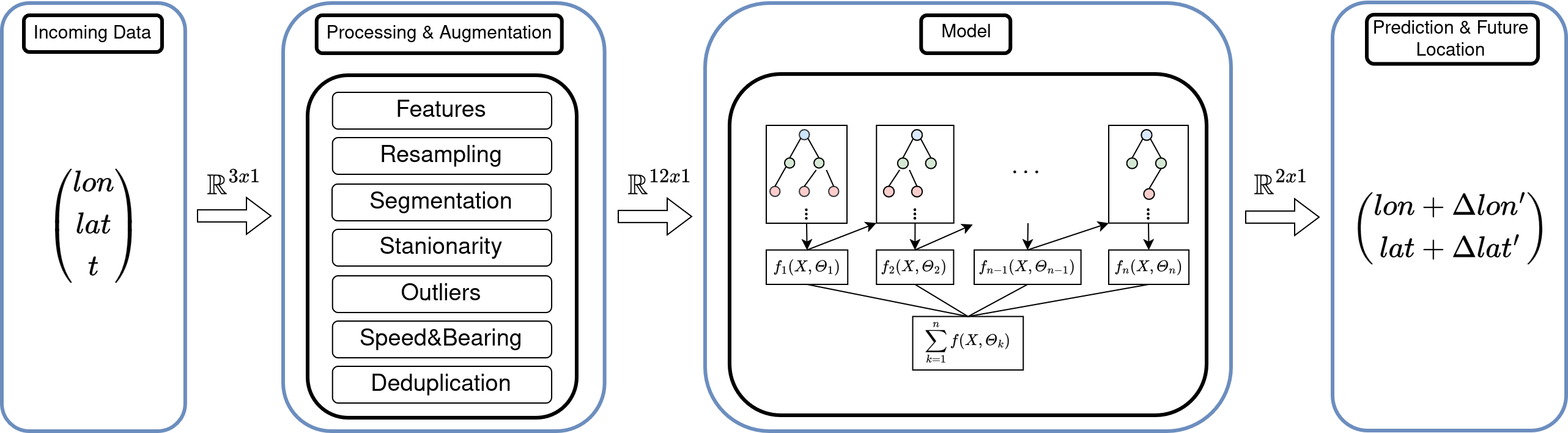}
    \caption{The proposed FLP-XR methodology.}
    \label{fig:model_arch}
    \vspace{-8pt}
\end{figure*}

\section{FLP-XR: Our Approach for Efficient Resource-Constrained FLP}\label{sec:III}

\subsection{Problem Formulation}
Given a stream of records $[p(1),...,p(t)]$ generated by $N$ vessels $V_i$, where each record $V_i(t)$ contains the vessel's unique identifier ($ID$), coordinates ($lon(t), lat(t)$) and the timestamp ($t_i$), the problem at hand is to train a Machine Learning (ML) model able to estimate the record $V_i(t+\Delta t)$, corresponding to $\Delta t$ time horizon in the future:

\begin{equation}
    V_i(t+\Delta_t) = [p_{lon}(t+\Delta t), p_{lat}(t+\Delta t)]
    \label{eq:flp}
\end{equation}

It is obvious that the time horizon $\Delta t$ clearly plays a significant role in the accuracy of the model's predictions. In our experimental study (Section \ref{sec:IV}), we examine $\Delta t$ at intervals of 10 minutes, up to a prediction horizon of 60 minutes.

\subsection{The proposed architecture} \label{subsec:III-B}
As the backbone of our prediction mechanism, we employ XGBoost\cite{xgboost}, a powerful ML algorithm based on the gradient boosting framework. The algorithm works by building multiple decision trees sequentially, where each new tree aims to correct the mistakes of the previous ones. It starts by making a single tree to predict the data, then analyzes where it went wrong, focusing particularly on instances it predicted poorly. Gradient descent is used to optimize the corrections made by each tree, minimizing errors in a way that adds up to a highly accurate model over many trees. Also, it is particularly effective in handling large datasets, missing values, and complex data patterns due to its regularization techniques, which reduce overfitting, and its parallelized implementation, which speeds up training.

Let us recall that the primary objective of our FLP-XR model is to optimize adaptability and performance, in other words create a prediction model capable of delivering accurate predictions across multiple time horizons using a single model instance. This removes the need to deploy multiple separately trained models, as it is the case in e.g., \cite{Eva}, \cite{tritsarolis2024}, thus improving computational efficiency and ease of use. To achieve this, we define the "target" prediction for each record as the coordinates of the vessel at a future time point, denoted as $\Delta t$ minutes in the future, where $\Delta t$ is the prediction horizon value. Additionally, by using multiple horizon values for each record, we gain the flexibility to predict positions across a range of future intervals. Because XGBoost operates exclusively on tabular data, we reformulate the prediction task so that each pair of known coordinates at time $t$ maps to the subsequent pair of predicted coordinates at time $t + \Delta t$. In summary, our proposed methodology, illustrated in Figure \ref{fig:model_arch}, begins by receiving the vessel’s trajectory in terms of coordinates and timestamps. We then apply various preprocessing and augmentation procedures (to be detailed in Section \ref{subsec:III-C}), including data deduplication, outlier and stationarity detection, resampling, segmentation, and feature extraction. Subsequently, an XGBoost model is employed to predict the vessel’s incremental changes in longitude and latitude ($\Delta lon$, $\Delta lat$), which are added to the most recent known position to obtain the vessel’s future location. 

\subsection{Preprocessing \& Feature Engineering}\label{subsec:III-C}

Transforming the available data sources into usable and feature-rich datasets that can be then fed into our model is a very important step of our proposed workflow. AIS data are inherently noisy, meaning that many of the existing features, like speed or bearing, that might come alongside the coordinates of a moving vessel might not be always trustworthy. For this reason, the main source of information that we built upon is the coordinated pair of each vessel at a specific timestamp. Based on these three values, i.e. $<lat(t), lon(t), t>$, we implement the following workflow in order to clean and augment the underlying dataset.

\subsubsection{Deduplication} 
We remove duplicate records that indicate that a single moving vessel appears in multiple places at identical timestamps; actually, the first one is kept the rest are discarded. 

\subsubsection{Speed and Bearing Calculation} 
Afterwards, we calculate the speed (in knots) and bearing (in degrees) of each vessel at timestamp $t_n$ using its coordinates at $t_{n-1}$ and $t_n$.

\begin{table}[!htbp]
\centering
\caption{Model Features}
\begin{center}
\begin{tabular}{ |c|c| } 
 \hline
 Feature name & Description \\[0.5ex] 
 \hline
 \hline
 \textit{v\_type} & vessel type  \\ 
 \hline
 \textit{\makecell{coords\\(lon, lat)}} & vessel location coordinates \\
 \hline
 \textit{sp} & vessel speed calculated at a given timestamp\\ 
 \hline
 \textit{br} & vessel bearing alculated at a given timestamp\\
 \hline
 \textit{\makecell{extrap\_diff\\(lon,lat)}} & \makecell{difference between the latest coordinates\\ and future coordinates assuming linear extrapolation\\ ($\Delta t$ amount of time later)} \\
 \hline
\textit{\makecell{last\_diff\\(lon,lat)}} & \makecell{difference between the latest coordinates \\and past coordinates ($\Delta t$ amount of time earlier)} \\ 
 \hline
 \textit{origin} & \makecell{the POI closest to the starting point of a given\\vessel's trip}  \\ 
 \hline
 \textit{orig\_dist} & \makecell{distance between the latest location and the starting\\ point of a given
vessel’s trip} \\
 \hline
 \textit{$\Delta t$} & prediction time horizon \\
 \hline
\end{tabular}
\end{center}
\label{tab:features}
\vspace{-25pt}
\end{table}

\subsubsection{Outlier Detection and Removal} 
Data points that deviate significantly from the expected distribution can introduce noise and bias in model training leading to overfitting or underfitting and thereby impairing the model's predictive performance. In our case, we focus on detecting and discarding speed-based outliers, by utilizing a predefined threshold ($s_{max}$). Applying more complex outlier detection methods that exist in the literature is beyond the scope of this work.

\subsubsection{Stationarity Detection and Removal} 
Based on the calculated speed, we are also able to perform stationarity detection using a predefined speed threshold ($s_{min}$). By detecting and discarding records of stationary vessels we result in saving on memory and improving performance. 

\subsubsection{Trip Segmentation}
Splitting the overall series of recorded locations of a vessel into smaller and more meaningful segments corresponding to trips is a key part of our preprocessing workflow. There are two criteria for trip segmentation, a temporal and spatial one. The temporal one relates to the fact that trying to estimate the future location of a vessel that has not transmitted its location for long time is risky. As such, we set a temporal threshold, $gap_{max}$, between the timestamps of two consecutive points, and when detected, these temporal gaps signal the completion of a trip and the beginning of a new one. On the other hand, the spatial criterion segments trajectories when a vessel stops (i.e., its calculated speed is less than $s_{min}$). After the segmentation, a trip is considered insignificant and is removed if it consists of less than $length_{min}$ points. For the remaining trips, we perform the following context augmentation: if the first point of the trip is within a threshold distance, $d_{min}$, of a Point of Interest (POI), the unique ID of that POI is assigned as the trip’s origin.

\subsubsection{Fixed Resampling}
In the mobility domain, a variable sampling rate of the recorded locations of moving objects poses significant challenges to ML models, especially in RNN architectures like LSTM. As discussed in \cite{Eva}\cite{Philip}, irregular sampling can lead to inconsistent temporal intervals, making it difficult for the network to learn temporal dependencies accurately and thus greatly affecting the model's predictive ability. In our model, in order to ensure consistency in the overall sampling rate of the dataset, we employ linear resampling using a given $rate$ value.

\subsubsection{Feature Selection} 
To maintain efficiency, we emphasize feature simplicity, selecting features that provide strong predictive power while minimizing computational overhead. By avoiding complex transformations or feature engineering steps that would slow down preprocessing, we ensure the model’s suitability for real-time applications, optimizing both the latency and accuracy that are essential for FLP. Based on the aforementioned preprocessing steps, for each incoming triple $<lat(t), lon(t), t>$ we extract 12 features, as they are listed in Table \ref{tab:features}.

Finally, we prepare our model to handle different prediction horizons, $\Delta t$, $\Delta t'$, $\Delta t''$, etc., by constructing the required source–target coordinate pairs (see Section \ref{subsec:III-B}). Specifically, for each triple $<lat(t), lon(t), t>$, we create different variations, $<lat(t+\Delta t), lon(t+\Delta t), t+\Delta t>$, $<lat(t+\Delta t'), lon(t+\Delta t'), t+\Delta t'>$, $<lat(t+\Delta t''), lon(t+\Delta t''), t+\Delta t''>$, etc., corresponding to the respective prediction horizons we aim to support.

\begin{figure*}
    \centering
    \subfigure[]
    {
        \includegraphics[width=2in]{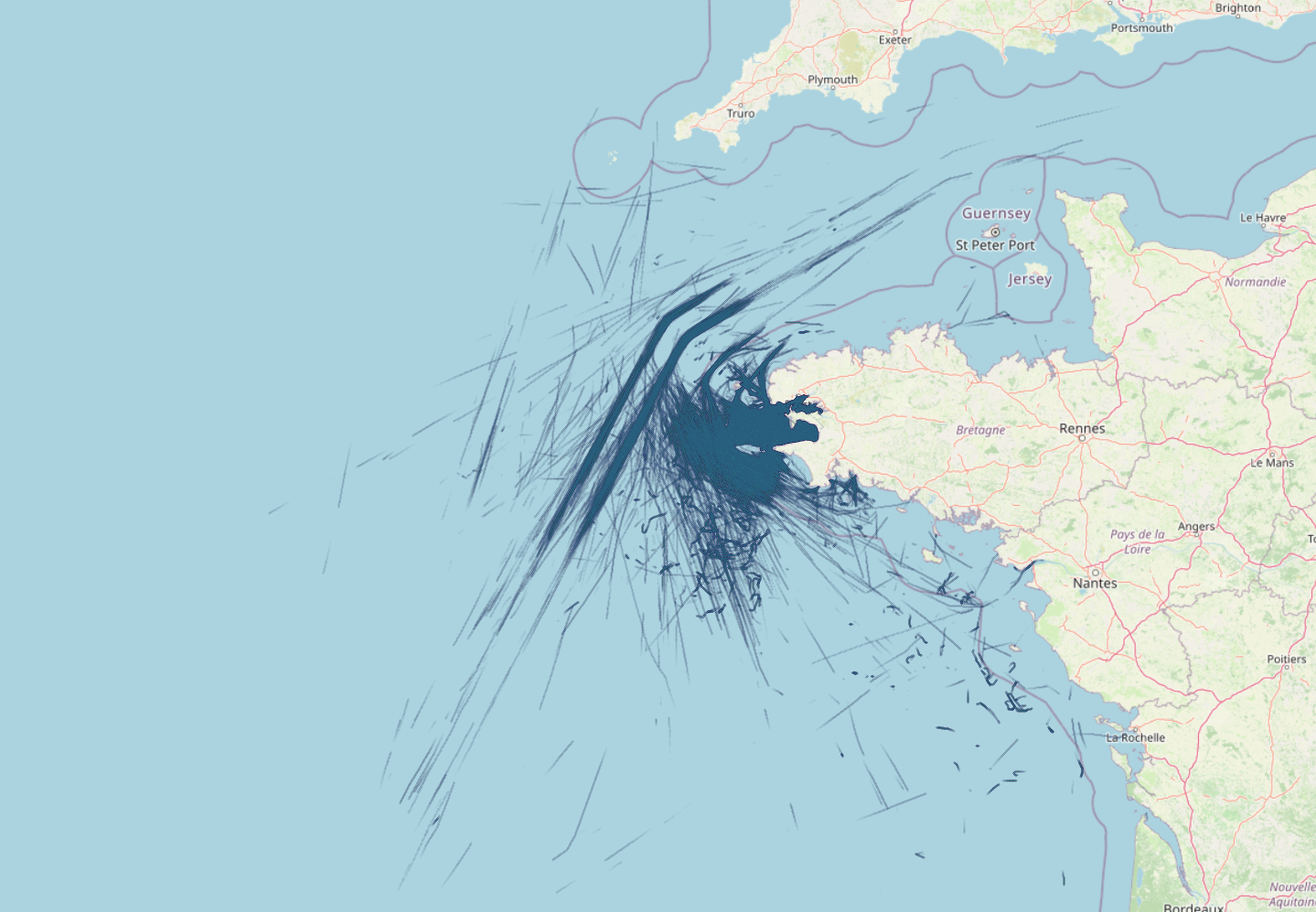}
        \label{fig:brest_fig}
    }
    \subfigure[]
    {
        \includegraphics[width=2in]{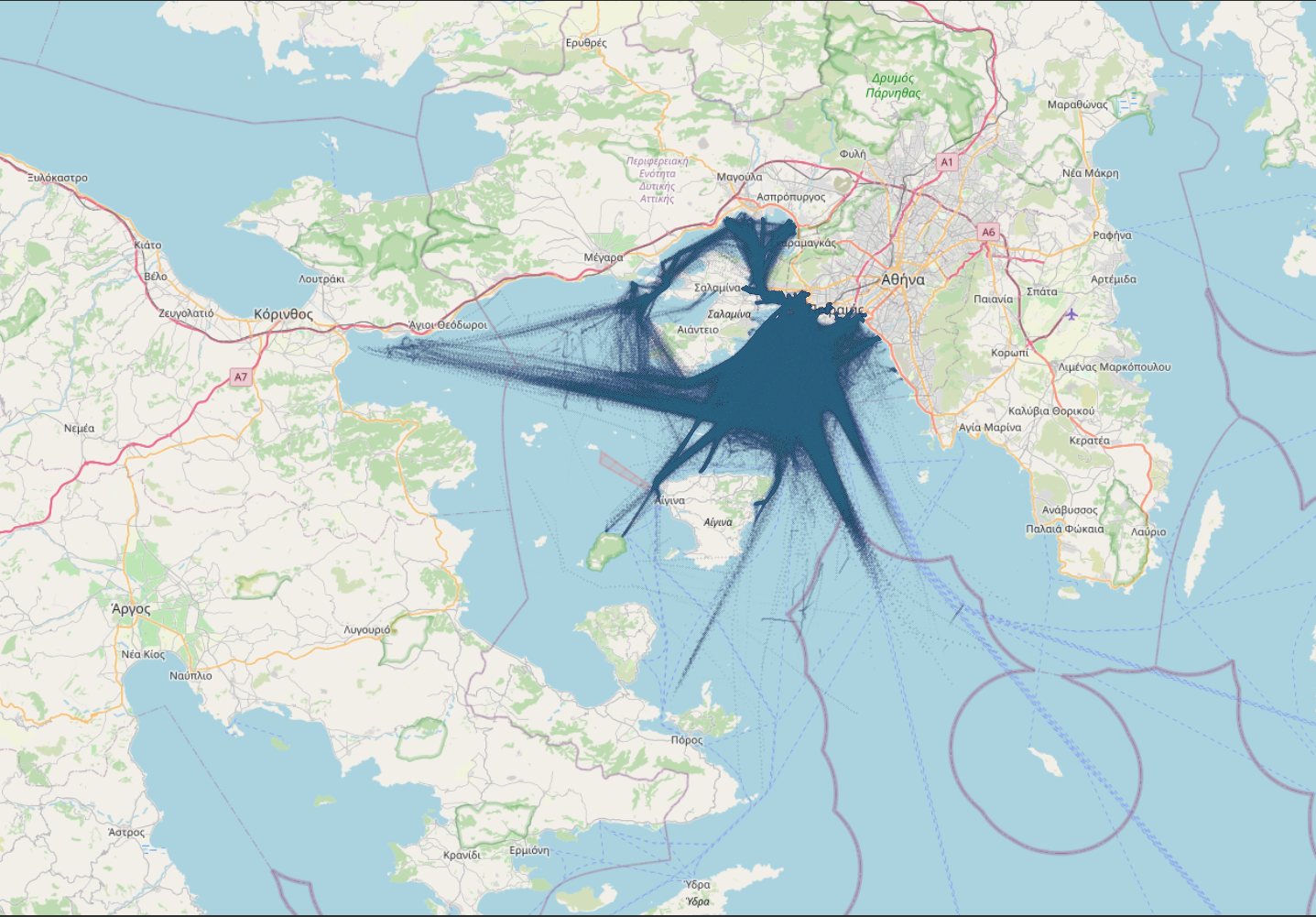}
        \label{fig:piraeus_fig}
    }
     \subfigure[]
    {
        \includegraphics[width=2in]{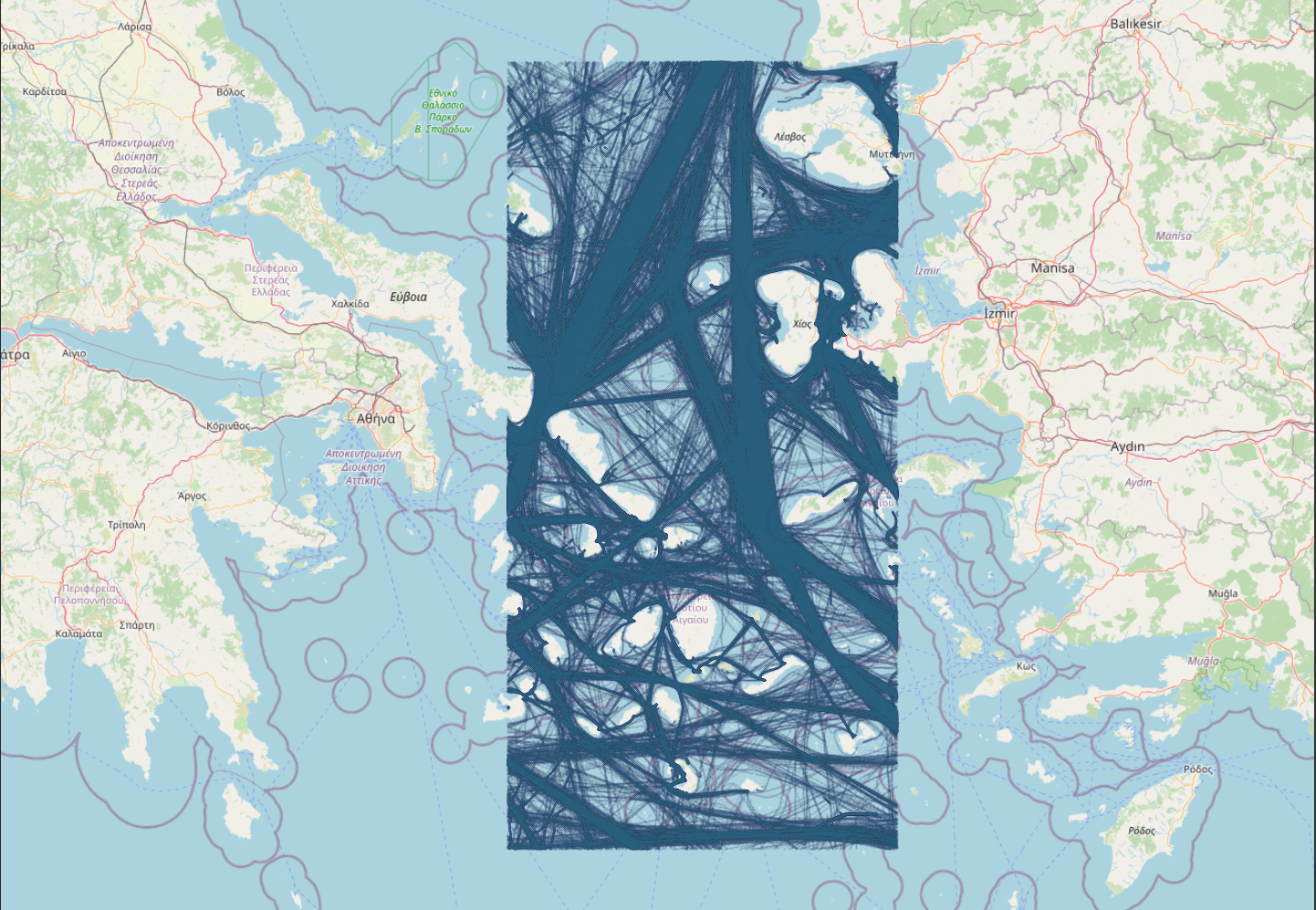}
        \label{fig:aegean_fig}
    }
    \caption{Snapshots of datasets used in our experimental study: (a) Brest; (b) Piraeus; (c) Aegean.}
    \label{fig:datasets_viz}
\end{figure*}

\begin{table*}
\caption{Statistics of the processed datasets for the purposes of model training.}
\centering
\begin{tabular}{llccccccc}
\toprule
 &  & \multicolumn{6}{c}{Model input for prediction horizon ($\Delta t$)} \\
\cmidrule(lr){3-8}
Dataset & Statistics & 10min. & 20min. & 30min. & 40min. & 50min. & 60min. \\
\midrule
\multirow{3}{*}{Brest} 
   & \#Vessels & 1644   & 1541   & 1453   & 1388   & 1300   & 1187   \\ 
   & \#Trips & 8966   & 7750   & 6828   & 6214   & 5389   & 4360   \\ 
   & \#Points & 746255 & 604482 & 490523 & 408163 & 325414 & 257079 \\ 
\midrule
\multirow{3}{*}{Piraeus} 
   & \#Vessels & 2568   & 2205   & 1833   & 1563   & 1097   & 674    \\ 
   & \#Trips & 17676  & 12034  & 7587   & 5507   & 3159   & 1634   \\ 
   & \#Points & 743637 & 423124 & 230021 & 138307 & 73809  & 41054  \\ 
\midrule
\multirow{3}{*}{Aegean} 
   & \#Vessels & 2918   & 2908   & 2900   & 2891   & 2879   & 2853   \\ 
   & \#Trips & 9657   & 9196   & 8860   & 8562   & 8062   & 7432   \\ 
   & \#Points & 2155218& 2044590& 1943762& 1862383& 1773266& 1692399\\ 
\bottomrule
\end{tabular}
\label{tab:Brest_dataset}
\vspace{-15pt}
\end{table*}

\section{Experimental Study} \label{sec:IV}
In this section, we evaluate the performance of our FLP-XR model on a variety of hardware, from high-performance computational systems to resource-constrained devices, using three real-world maritime mobility datasets, and present our experimental results.

\subsection{Experimental Setup, Datasets and Preprocessing} \label{subsec:V-A}

In this experimental study, we utilize two popular open AIS datasets, namely "Brest"\footnote{The dataset is available at \href{https://doi.org/10.5281/zenodo.1167595}{https://doi.org/10.5281/zenodo.1167595}}, "Piraeus"\footnote{The dataset is available at \href{https://doi.org/10.5281/zenodo.6323416}{https://doi.org/10.5281/zenodo.6323416}} as well as one closed dataset called "Aegean"\footnote{The dataset has been kindly provided by Kpler for research
purposes, in the context of the EU project MobiSpaces at \href{https://mobispaces.eu}{https://mobispaces.eu}}.

The Brest dataset\cite{ray_dataset_brest} is a collection of vessel movements that spans a
six-month period from October 1st, 2015, to March 31st, 2016, and includes vessel
positions over the Celtic Sea, the North Atlantic Ocean, the English Channel, and the Bay of Biscay. The dataset is structured into four categories: navigation data, vessel-oriented data, geographic data, and environmental data. In our study, we utilise the entire 6-months dataset, consisting of 19,035,631 AIS records, alongside port and vessel type information.

The Piraeus dataset \cite{tritsarolis_dataset_piraeus}, is a rich repository of maritime data, encompassing over 244 million AIS records collected over a period of more than two and a half years, from May 9th, 2017, to December 26th, 2019. This dataset includes not only (anonymized) vessel positions but also correlated data, such as weather conditions, offering a valuable resource for research and analysis in maritime logistics, traffic patterns, and safety studies. In our study, we utilise an 1-year subset of the dataset (from January 1st to December 31st, 2019), 92,534,304 records in total, alongside port and vessel type information.

Finally, the Aegean dataset is a collection of vessel movements across the Aegean sea, spanning the single-month period of November of 2018, consisting of 7,352,408 records, alongside port of vessel type information.  
Figure \ref{fig:datasets_viz} illustrates these datasets on the map.

\begin{table*}
\caption{Prediction error ($\mu \pm \sigma$) of FLP-XR VS. Nautilus per dataset (less is better).}
\centering
\begin{tabular}{cccccccc}
\toprule
 &  & \multicolumn{6}{c}{Prediction horizon ($\Delta t$)} \\
\cmidrule(lr){3-8}
Dataset & Model & 10min. & 20min. & 30min. & 40min. & 50min. & 60min. \\
\midrule
\multirow{2}{*}{Brest} & FLP-XR & \textbf{429}$\pm$544 & 1066$\pm$1178 & \textbf{1749}$\pm$1796 & 2364$\pm$2470 & 3119$\pm$3215 & 3935$\pm$3986 \\
& Nautilus\cite{tritsarolis2024} & 509$\pm$806 & \textbf{988}$\pm$1379 & 2112$\pm$2489 & \textbf{2197}$\pm$2349 & \textbf{2798}$\pm$2705 & \textbf{3912}$\pm$3770 \\
\midrule
\multirow{2}{*}{Piraeus} & FLP-XR & 339$\pm$369 & 838$\pm$811 & 1374$\pm$1281 & 1874$\pm$1759 & 2598$\pm$2378 & \textbf{3334}$\pm$2852\\
& Nautilus\cite{tritsarolis2024} & \textbf{234}$\pm$501 & \textbf{492}$\pm$1078 & \textbf{442}$\pm$1104 & \textbf{831}$\pm$2020 & \textbf{1475}$\pm$3283 & 3818$\pm$5668\\
\midrule
\multirow{2}{*}{Aegean} & FLP-XR & \textbf{176}$\pm$615 & \textbf{460}$\pm$1040 & \textbf{809}$\pm$1508 & \textbf{1123}$\pm$1914 & \textbf{1543}$\pm$2488 & 2003$\pm$3027\\
& Nautilus\cite{tritsarolis2024} & 264$\pm$392 & 615$\pm$607 & 1205$\pm$1720 & 1563$\pm$1795 & 3246$\pm$4022 & \textbf{1978}$\pm$1566 \\
\bottomrule
\end{tabular}
\label{tab:quality_eval}
\vspace{-15pt}
\end{table*}

In accordance with the preprocessing pipeline outlined in Section \ref{subsec:III-C}, all datasets underwent a comprehensive cleaning and preparation process, by applying the following parameter values: $s_{min}=1$ knot, $s_{max}=50$ knots, $gap_{max}=45$ minutes, $length_{min}=30$ points, and $d_{min}=1$ nautical mile. 
As for the resampling step, we experimented with different sampling rate values, from $30$ to $150$ seconds, and we selected $rate=90$ sec for the final adjustment of the trajectories, as this configuration produced the best results. A detailed account of these experiments is provided in Tables \ref{tab:grid_search_brest}, \ref{tab:grid_search_piraeus} and \ref{tab:grid_search_aegean}. Statistics of the processed datasets, ready to train our model for six different prediction horizons, from 10 minutes to 60 minutes, are presented in Table \ref{tab:Brest_dataset}.

In the section that follows, we provide the experimental results of our study, comparing the prediction performed by our XGBoost based model with the current SotA\cite{tritsarolis2024}. To evaluate the quality of the predictions, we will calculate the displacement error between predicted and target points at identical timestamps. Given that we require this error to be interpretable in real-world terms and that our datasets are provided in the EPSG:4326 coordinate system (latitude and longitude), for the purpose of error calculation we utilize the Haversine distance (in meters).

The entire preprocessing pipeline (see Section \ref{subsec:III-C}) was implemented in Rust; the justification of this decision will be discussed in Section \ref{subsec:IV-D}. The FLP-XR model was implemented in Python using the  XGBoost\footnote{Python library utilized for model implementation \href{https://xgboost.readthedocs.io/en/stable/}{https://xgboost.readthedocs.io}} library. Our FLP-XR model running times were calculated on three different hardware environments: (i) a computing cluster featuring an Intel Xeon x86 CPU, a single Nvidia A100 GPU and 1TB of RAM (ii) an Intel NUC with an i7 x86 CPU and 16 GB of RAM, and (iii) a Raspberry Pi 4 (RPi 4) with an ARM CPU and 8 GB of RAM. On the other hand, the SotA model\cite{tritsarolis2024} was performed in hardware setting (i) above, since it is not transferable to resource-constrained environments, such as (ii) and (iii) above. For reproducibility purposes, the source code used in our experimental study is publicly available at GitHub\footnote{\href{https://github.com/DataStories-UniPi/FLP-XR}{https://github.com/DataStories-UniPi/FLP-XR}}.

\subsection{XGBoost hyperparameter tuning}
Given our objective to construct a highly accurate and computationally efficient FLP model, the selection of appropriate hyperparameters is of critical importance. The key hyperparameters selected for tuning include \textit{tree\_method}, \textit{max\_depth}, \textit{learning\_rate}, and \textit{n\_estimators}. To optimize the performance of the XGBoost model, a grid search was conducted to identify the best hyperparameter values. The optimal values were selected based on the configuration that yielded the highest performance according to the chosen evaluation metric, ensuring the model achieved a good balance between accuracy and generalizability (see Tables \ref{tab:grid_search_brest}, \ref{tab:grid_search_piraeus} and \ref{tab:grid_search_aegean}).
In particular, the \textit{tree\_method} parameter was set to "hist" which is a variant that constructs histograms for fast and memory-efficient training, particularly beneficial for large datasets. 
The \textit{max\_depth} parameter, set to 12, controls the maximum depth of each tree within the ensemble model. By setting this depth, the model captures complex patterns in the data while helping to prevent overfitting by limiting the extent of feature interaction. A \textit{learning\_rate} of 0.01 was chosen to regulate the step size during each update, balancing between model convergence speed and generalization.
Finally, we set \textit{n\_estimators} to 750, which specifies the number of boosting rounds or trees the model builds iteratively.

\subsection{Performance Comparison with SotA}
Before presenting our experimental results, it is important to highlight a key difference in how the two models—our proposed model and the SotA\cite{tritsarolis2024}—handle prediction horizons. Specifically, our model can generate predictions at exact horizon time (after precisely $\Delta t$ minutes), whereas the Nautilus\cite{tritsarolis2024} model can only produce predictions at the level of horizon intervals (e.g., after (5, 10] minutes), as it does not rely on a dataset with a fixed sampling rate. To ensure a fair comparison, Nautilus predictions reported for horizon $\Delta t$ correspond to the average of two values: the prediction at horizon interval $(\Delta t-k]$ and the prediction at horizon interval $(\Delta t+k]$, where $k=5$ min. is the step of the prediction horizons. For instance, to get the prediction error of Nautilus that corresponds to horizon $\Delta t=10$ of FLP-XR, we take the average of prediction buckets (5,10] and (10,15]. The only exception is the final prediction (60 minutes), where Nautilus prediction corresponds to bucket (55, 60], since no subsequent predictions exist.

Each of the processed datasets (see Table \ref{tab:Brest_dataset}) was partitioned into 80\% training and 20\% testing sets. Given the time-dependent nature of the data, this splitting was conducted in chronological order, according to the timestamps associated with the vessel locations.

For quality evaluation, Table \ref{tab:quality_eval} details the displacement error, in meters, across various prediction horizons, from 10 to 60 minutes, providing a comparative analysis between our FLP-XR model and Nautilus\cite{tritsarolis2024}.
FLP-XR demonstrates superior performance over Nautilus for certain prediction horizons, while underperforming in others. On the Brest dataset, FLP-XR reduces the error by around 15\% with respect to SotA for 10 and 30-minute horizons, while errors for other horizons remain closely aligned, with SotA slightly outperforming our proposal by 1\% - 10\%. Notably, the error progression of FLP-XR is more stable across all horizons, increasing linearly with the prediction interval. In contrast, Nautilus displays inconsistent error progression, with pronounced fluctuations, such as a sharp increase in the $[30, 40]$ minute range. These irregularities in Nautilus may stem from limitations in the bucket sizes used for prediction, potentially due to insufficient training data in certain ranges.
On the Piraeus dataset, FLP-XR shows inferior performance, except for the 60-minute horizon where it outperforms Nautilus by 13\%. As with the Brest dataset, FLP-XR maintains a linear increase in prediction error with the horizon, while Nautilus exhibits error fluctuations. For example, the Nautilus error reported at the 30-minute horizon violates the linear increase trend that one might expect to see; this may be attributed to data sparsity in these prediction buckets.
On the Aegean dataset, FLP-XR outperforms the SotA model in all horizons except for the 60-minute prediction interval. Consistent with observations from the other datasets, FLP-XR shows a linear trend in error progression across horizons. In contrast, Nautilus demonstrates irregular increases and decreases in error, coupled with very high standard deviations in certain cases, such as the 50-minute horizon. These claims are further supported by examining the standard deviation of errors: FLP-XR consistently exhibits standard deviations close to the mean, while Nautilus shows significantly higher standard deviations, up to double or triple in some cases, indicating less predictable and less consistent predictions. Overall, these patterns highlight the robustness and stability of FLP-XR in comparison to the SotA model, particularly in handling datasets characterized by high variability.

Beyond the error metrics, performance in both training and inference times is another critical aspect of our model’s value proposition. As illustrated in Table \ref{tab:efficiency_eval}, FLP-XR demonstrates significant speed advantages over SotA, with training and inference time improvements being at the level of two and three orders of magnitude, respectively, calculated at the same (Xeon+A100) hardware environment; compare the first and the fourth row in Table \ref{tab:efficiency_eval}. Specifically, training time improvement achieved by FLP-XR ranges from 170x (Aegean) to 520x (Brest). Similar improvements are observed in inference times, where FLP-XR achieves substantial speedups, ranging from 2175x (Piraeus) to 5035x (Aegean). This remarkable difference in runtime translates into the conclusion that FLP-XR is able to perform over 1000 predictions per second, in contrast to SotA’s capability of generating on the average less than 1 prediction within the same time frame.

\begin{table}
\caption{Comparison of training and inference(per AIS message) times between FLP-XR and related work (less is better).}
\centering
\begin{tabular}{cccc}
\toprule
Dataset & Model & Training($sec$) & Inference($\mu sec$) \\
\midrule
\multirow{4}{*}{Brest} & FLP-XR(Xeon+A100) & 69 & 0.56 \\
 & FLP-XR(NUC-i7) & 186 & 12 \\
 & FLP-XR(RPi 4) & 1119 & 118 \\
 & Nautilus(Xeon+A100)\cite{tritsarolis2024} & 35909 & 1345 \\
\midrule
\multirow{4}{*}{Piraeus} & FLP-XR(Xeon+A100) & 60 & 0.61 \\
 & FLP-XR(NUC-i7) & 123 & 14 \\
 & FLP-XR(RPi 4) & 811 & 121 \\
 & Nautilus(Xeon+A100)\cite{tritsarolis2024} & 13161 & 1327 \\
\midrule
\multirow{4}{*}{Aegean} & FLP-XR(Xeon+A100) & 86 & 0.28 \\
 & FLP-XR(NUC-i7) & 643 & 10 \\
 & FLP-XR(RPi 4) & 4020 & 105 \\
 & Nautilus(Xeon+A100)\cite{tritsarolis2024} & 14637 & 1410 \\
\bottomrule
\end{tabular}
\label{tab:efficiency_eval}
\vspace{-15pt}
\end{table}

\subsection{Performance in Resource-Constrained Environments} \label{subsec:IV-D}
By leveraging an efficient ML algorithm (XGBoost) and adopting a scalable tabular data model rather than the most common time-series-based approach, we have developed a resource-efficient solution that can be readily adapted for diverse cloud computing environments, including Edge and Fog platforms. Current state-of-the-art solutions, which often rely on RNN architectures\cite{Eva},\cite{yuhao_li24},\cite{tritsarolis2024}, demand significant computational resources and are largely impractical for deployment in environments with limited hardware capabilities. These architectures typically require long inference times and extensive training on powerful GPU clusters, limiting their feasibility in real-time, edge-based applications. In response to these challenges, we developed an XGBoost-based solution tailored for CC environments, ensuring it meets strict performance and resource requirements while maintaining or exceeding the accuracy of existing solutions.

\begin{table}
\caption{FLP-XR average throughput.}
\centering
\begin{tabular}{cccccc}
\toprule
Dataset & Hardware & \makecell{Batch size\\(\#records)} & \makecell{Preprocessing\\per record\\($\mu sec$)} & \makecell{Inference\\per record\\($\mu sec$)} & \makecell{Throughput \\per batch\\(sec)} \\
\midrule
\multirow{6}{*}{Brest} 
 & \multirow{2}{*}{\makecell{Xeon+\\A100}} & 10K & 5.69 & 9.97 & 0.16 \\
 &                         & 100K & 4.37 & 1.17 & 0.55 \\
\cmidrule(lr){2-6}
 & \multirow{2}{*}{NUC-i7}   & 10K & 1.02 & 9.11 & 0.10 \\
 &                         & 100K & 0.80 & 9.01 & 0.98 \\
\cmidrule(lr){2-6}
 & \multirow{2}{*}{RPi 4}    & 10K & 4.96 & 102.04 & 1.07 \\
 &                         & 100K & 3.83 & 95.03 & 9.88 \\
\midrule
\multirow{6}{*}{Piraeus} 
 & \multirow{2}{*}{\makecell{Xeon+\\A100}} & 10K & 2.65 & 14.41 & 0.17 \\
 &                         & 100K & 2.98 & 1.35 & 0.43 \\
\cmidrule(lr){2-6}
 & \multirow{2}{*}{NUC-i7}   & 10K & 0.50 & 9.38 & 0.10 \\
 &                         & 100K & 0.51 & 9.17 & 0.97 \\
\cmidrule(lr){2-6}
 & \multirow{2}{*}{RPi 4}    & 10K & 2.02 & 74.52 & 0.77 \\
 &                         & 100K & 2.05 & 71.77 & 7.38 \\
\midrule
\multirow{6}{*}{Aegean} 
 & \multirow{2}{*}{\makecell{Xeon+\\A100}} & 10K & 100.52 & 16.41 & 1.17 \\
 &                         & 100K & 110.71 & 1.89 & 11.26 \\
\cmidrule(lr){2-6}
 & \multirow{2}{*}{NUC-i7}   & 10K & 24.03 & 8.29 & 0.32 \\
 &                         & 100K & 26.04 & 8.08 & 3.41 \\
\cmidrule(lr){2-6}
 & \multirow{2}{*}{RPi 4}    & 10K & 132.70 & 84.11 & 2.17 \\
 &                         & 100K & 144.64 & 74.96 & 21.96 \\
\bottomrule
\end{tabular}
\label{tab:throughput}
\vspace{-20pt}
\end{table}

To validate our claims, we conducted a series of experiments to evaluate the efficiency of the proposed FLP-XR in resource-constrained environments. As we observe in Table \ref{tab:efficiency_eval}, our proposed model achieves groundbreaking results, setting new benchmarks in both training and inference speeds. Remarkably, the model's training process on edge devices is approximately 12x faster than the state-of-the-art solution trained on a powerful computing cluster environment. This achievement underscores the efficiency of our approach in drastically reducing the computational and time overheads typically required for model development. Furthermore, the inference speed on edge devices is 24x faster compared to the SotA running inference on GPU, demonstrating the unparalleled efficiency and suitability of our model for real-time applications in resource-constrained environments. 

Finally, to optimize the performance of our architecture, as we mentioned earlier we implemented the entire preprocessing pipeline in Rust, selected for its strong performance across a wide range of computing environments. In order to demonstrate it, we perform the following experiment: for each combination of dataset and hardware environment and having already processed the half of the dataset (for fairness purposes, in order to avoid cold start issues), we feed FLP-XR (see Figure \ref{fig:model_arch}) with a batch of records (in two alternative sizes, either 10K or 100K records per batch) and we calculate the time required per record for the preprocessing and the inference step. In Table \ref{tab:throughput}, apart from the respective times required for each of the two steps per record (in microseconds), we calculate the end-to-end throughput of our pipeline for the entire batch (in seconds), as the sum of the preprocessing and inference times multiplied by the number of records in the batch. It is clear from these figures that the proposed architecture is able to support a fleet of order 100K (see Xeon+A100 and NUC-i7 rows) or 10K (see RPi 4 rows) vessels in usually less than one or at most few seconds per batch. To put this outcome into a real use-case perspective, in a cloud environment, an AIS monitoring system with world-wide coverage tracks approximately 270K vessels concurrently\footnote{Indicatively, see the statistics of the MarineTraffic platform (\href{https://www.marinetraffic.com}{https://www.marinetraffic.com}).}, which transmit their position every few seconds while underway\footnote{The current AIS protocol guidelines define a 2 to 10 seconds transmission rate of Class A AIS positions while a vessel is underway (and every 3 minutes while at anchor); source: \href{https://www.navcen.uscg.gov/ais-class-a-reports}{https://www.navcen.uscg.gov/ais-class-a-reports}.}, whereas in an edge environment, an AIS antenna located in a busy port could potentially track around 2K vessels\footnote{Indicatively, see the statistics about the vessel density in Rotterdam area (\href{https://www.marinetraffic.com/en/details/areas/areaId:12/}{https://www.marinetraffic.com/en/details/areas/areaId:12/}) and Singapore area (\href{https://www.marinetraffic.com/en/details/areas/areaId:112}{https://www.marinetraffic.com/en/details/areas/areaId:112}).}. As it turns out from the findings of Table \ref{tab:throughput}, with the appropriate selection of hardware settings, our methodology is expected to be capable of providing real-time support under all these specs in the Computing Continuum. 

\section{Conclusion} \label{sec:V}
In this work, we introduced FLP-XR, a lightweight and efficient XGBoost-based model for future location prediction in the maritime domain, specifically targeting resource-constrained computing environments. While the model's prediction accuracy is slightly below the SotA in some cases, it demonstrates competitive performance and surpasses it in certain scenarios. Most importantly, it offers unprecedented advantages in terms of computational efficiency, achieving training speeds that are 300x faster and inference speeds that are 3000x faster than existing solutions.

The proposed model's lightweight nature and ability to function seamlessly in CC environments validate its suitability for real-world applications, particularly in scenarios where computational resources are limited, and low-latency predictions are critical. This work highlights the potential of non-RNN approaches in solving complex prediction tasks in resource-constrained domains, providing a new perspective on efficient, scalable, and deployable ML solutions.

Future work includes further feature engineering to enhance accuracy while retaining efficiency and conducting extensive evaluations on more and diverse maritime datasets to ensure robustness. We additionally aim to explore the adaptation of the model to the aviation domain, utilizing the ADS-B aircraft position broadcasting protocol, where similar resource-constrained environments and low-latency requirements exist.

\section{Acknowledgements} \label{sec:VI}
This work was supported in part by the Horizon Framework
Programme of the European Union under grant agreement No.
101070279 (MobiSpaces; \href{https://mobispaces.eu}{https://mobispaces.eu}). In this work,
Kpler provided the Aegean AIS dataset and the requirements
of the business case.

\bibliographystyle{plain}
\bibliography{bibliography}


\begin{table*}

\centering
\caption{Brest dataset \\base model: SR=90, LR=0.01, MaxDepth=12, Boosters=750}
\begin{subtable}
\centering
\begin{tabular}{llrrrllllll}
\toprule
 &  & Model size[mb] & Inference[$\mu$s] & Training[s] & 10 & 20 & 30 & 40 & 50 & 60 \\
Model & SR &  &  &  &  &  &  &  &  &  \\
\midrule
\multirow[c]{5}{*}{FLP-XR} & 30 & 294 & 0.33 & 89 & 481 & 1091 & 1732 & 2415 & 3143 & 3937 \\
 & 60 & 280 & 0.52 & 76 & 484 & 1099 & 1748 & 2424 & 3139 & \textbf{3936} \\
 & 90 & 270 & 0.56 & 69 & \textbf{428} & \textbf{1066} & 1749 & \textbf{2367} & \textbf{3110} & \textbf{3936} \\
 & 120 & 258 & 0.48 & 63 & 491 & 1103 & 1767 & 2479 & 3216 & 4022 \\
 & 150 & 251 & 1.01 & 59 & 484 & 1088 & \textbf{1745} & 2460 & 3233 & 4032 \\
\cline{1-11}
\bottomrule
\end{tabular}
\end{subtable}

\begin{subtable}
\centering
        \begin{tabular}{llrrrllllll}
\toprule
 &  & Model size[mb] & Inference[$\mu$s] & Training[s] & 10 & 20 & 30 & 40 & 50 & 60 \\
Model & Max Depth &  &  &  &  &  &  &  &  &  \\
\midrule
\multirow[c]{5}{*}{FLP-XR} & 6 & 7 & 0.31 & 9 & 512 & 1252 & 2032 & 2696 & 3516 & 4439 \\
 & 9 & 45 & 0.30 & 20 & 464 & 1130 & 1836 & 2450 & 3210 & 4037 \\
 & 12 & 270 & 0.56 & 69 & 428 & 1066 & 1749 & \textbf{2367} & \textbf{3110} & \textbf{3936} \\
 & 15 & 1174 & 0.99 & 273 & 409 & \textbf{1046} & \textbf{1738} & 2379 & 3151 & 4024 \\
 & 18 & 3546 & 3.04 & 858 & \textbf{404} & 1055 & 1771 & 2422 & 3205 & 4097 \\
\cline{1-11}
\bottomrule
\end{tabular}
\end{subtable}

\begin{subtable}
\centering
        \begin{tabular}{llrrrllllll}
\toprule
 &  & Model size[mb] & Inference[$\mu$s] & Training[s] & 10 & 20 & 30 & 40 & 50 & 60 \\
Model & LR &  &  &  &  &  &  &  &  &  \\
\midrule
\multirow[c]{5}{*}{FLP-XR} & 0.001 & 293 & 0.51 & 74 & 1265 & 2665 & 4041 & 5122 & 6382 & 7572 \\
 & 0.005 & 301 & 0.63 & 74 & 464 & 1144 & 1856 & 2472 & 3235 & 4082 \\
 & 0.01 & 270 & 0.56 & 69 & \textbf{428} & 1066 & 1749 & 2367 & \textbf{3110} & 3936 \\
 & 0.05 & 216 & 0.54 & 54 & 429 & \textbf{1048} & \textbf{1732} & \textbf{2357} & 3113 & \textbf{3931} \\
 & 0.1 & 207 & 0.43 & 53 & 444 & 1069 & 1762 & 2395 & 3156 & 3977 \\
\cline{1-11}
\bottomrule
\end{tabular}
\end{subtable}

\begin{subtable}
\centering
        \begin{tabular}{llrrrllllll}
\toprule
 &  & Model size[mb] & Inference[$\mu$s] & Training[s] & 10 & 20 & 30 & 40 & 50 & 60 \\
Model & Boosters &  &  &  &  &  &  &  &  &  \\
\midrule
\multirow[c]{5}{*}{FLP-XR} & 500 & 195 & 0.38 & 48 & 440 & 1094 & 1788 & 2404 & 3157 & 3995 \\
 & 625 & 234 & 0.58 & 58 & 432 & 1076 & 1763 & 2380 & 3125 & 3955 \\
 & 750 & 270 & 0.56 & 69 & 428 & 1066 & 1749 & 2367 & 3110 & 3936 \\
 & 875 & 303 & 0.57 & 75 & 428 & 1061 & 1744 & 2362 & 3106 & 3928 \\
 & 1000 & 339 & 0.72 & 84 & \textbf{427} & \textbf{1057} & \textbf{1739} & \textbf{2357} & \textbf{3101} & \textbf{3918} \\
\cline{1-11}
\bottomrule
\end{tabular}
\end{subtable}
\label{tab:grid_search_brest}
\end{table*}


\begin{table*}
\centering
\caption{Piraeus dataset \\base model: SR=90, LR=0.01, MaxDepth=12, Boosters=750}
\begin{subtable}
\centering
        \begin{tabular}{llrrrllllll}
\toprule
 &  & Model size[mb] & Inference[$\mu$s] & Training[s] & 10 & 20 & 30 & 40 & 50 & 60 \\
Model & SR &  &  &  &  &  &  &  &  &  \\
\midrule
\multirow[c]{5}{*}{FLP-XR} & 30 & 284 & 0.36 & 78 & 394 & 848 & 1332 & 1902 & 2535 & 3148 \\
 & 60 & 270 & 0.61 & 67 & 390 & 850 & 1340 & 1905 & 2537 & 3214 \\
 & 90 & 250 & 0.61 & 60 & \textbf{339} & \textbf{838} & 1374 & \textbf{1874} & 2598 & 3334 \\
 & 120 & 233 & 0.83 & 55 & 385 & 855 & \textbf{1358} & 1908 & \textbf{2579} & \textbf{3198} \\
 & 150 & 214 & 2.13 & 50 & 384 & 852 & 1372 & 1974 & 2598 & 3297 \\
\cline{1-11}
\bottomrule
\end{tabular}
\end{subtable}

\begin{subtable}
\centering
        \begin{tabular}{llrrrllllll}
\toprule
 &  & Model size[mb] & Inference[$\mu$s] & Training[s] & 10 & 20 & 30 & 40 & 50 & 60 \\
Model & Max Depth &  &  &  &  &  &  &  &  &  \\
\midrule
\multirow[c]{5}{*}{FLP-XR} & 6 & 7 & 0.20 & 6 & 431 & 1026 & 1620 & 2112 & 2802 & 3615 \\
 & 9 & 44 & 0.31 & 15 & 373 & 899 & 1441 & 1915 & 2601 & 3326 \\
 & 12 & 250 & 0.61 & 60 & 339 & 838 & \textbf{1374} & \textbf{1874} & \textbf{2598} & \textbf{3334} \\
 & 15 & 969 & 1.55 & 225 & 323 & \textbf{820} & 1377 & 1902 & 2650 & 3397 \\
 & 18 & 2595 & 3.43 & 638 & \textbf{318} & 833 & 1408 & 1936 & 2682 & 3475 \\
\cline{1-11}
\bottomrule
\end{tabular}
\end{subtable}

\begin{subtable}
\centering
        \begin{tabular}{llrrrllllll}
\toprule
 &  & Model size[mb] & Inference[$\mu$s] & Training[s] & 10 & 20 & 30 & 40 & 50 & 60 \\
Model & LR &  &  &  &  &  &  &  &  &  \\
\midrule
\multirow[c]{5}{*}{FLP-XR} & 0.001 & 256 & 0.94 & 63 & 1102 & 2213 & 3115 & 3878 & 4795 & 5659 \\
 & 0.005 & 274 & 0.93 & 65 & 368 & 889 & 1436 & 1930 & 2647 & 3426 \\
 & 0.01 & 250 & 0.61 & 60 & 339 & 838 & \textbf{1374} & \textbf{1874} & \textbf{2598} & \textbf{3334} \\
 & 0.05 & 189 & 0.88 & 47 & \textbf{329} & \textbf{829} & 1384 & 1904 & 2638 & 3350 \\
 & 0.1 & 182 & 0.82 & 45 & 332 & 840 & 1411 & 1934 & 2671 & 3400 \\
\cline{1-11}
\bottomrule
\end{tabular}
\end{subtable}

\begin{subtable}
\centering
        \begin{tabular}{llrrrllllll}
\toprule
 &  & Model size[mb] & Inference[$\mu$s] & Training[s] & 10 & 20 & 30 & 40 & 50 & 60 \\
Model & Boosters &  &  &  &  &  &  &  &  &  \\
\midrule
\multirow[c]{5}{*}{FLP-XR} & 500 & 179 & 0.60 & 42 & 349 & 857 & 1396 & 1891 & 2612 & 3363 \\
 & 625 & 216 & 0.64 & 51 & 342 & 844 & 1380 & 1878 & 2601 & 3341 \\
 & 750 & 250 & 0.61 & 60 & 339 & 838 & 1374 & \textbf{1874} & \textbf{2598} & 3334 \\
 & 875 & 278 & 0.99 & 67 & 337 & 835 & 1372 & 1875 & 2600 & \textbf{3332} \\
 & 1000 & 305 & 0.81 & 74 & \textbf{336} & \textbf{833} & \textbf{1371} & \textbf{1874} & 2602 & 3335 \\
\cline{1-11}
\bottomrule
\end{tabular}
\end{subtable}
\label{tab:grid_search_piraeus}
\end{table*}


\begin{table*}
\centering
\caption{Aegean dataset \\base model: SR=90, LR=0.01, MaxDepth=12, Boosters=750}
\begin{subtable}
\centering
        \begin{tabular}{llrrrllllll}
\toprule
 &  & Model size[mb] & Inference[$\mu$s] & Training[s] & 10 & 20 & 30 & 40 & 50 & 60 \\
Model & SR &  &  &  &  &  &  &  &  &  \\
\midrule
\multirow[c]{5}{*}{FLP-XR} & 30 & 289 & 0.29 & 146 & 198 & 468 & 794 & 1149 & 1545 & 1994 \\
 & 60 & 281 & 0.27 & 101 & 201 & 473 & 799 & 1153 & 1544 & 1981 \\
 & 90 & 273 & 0.28 & 86 & \textbf{176} & \textbf{460} & \textbf{809} & \textbf{1123} & \textbf{1543} & \textbf{2003} \\
 & 120 & 264 & 0.27 & 79 & 208 & 485 & 813 & 1174 & 1570 & 2009 \\
 & 150 & 258 & 0.29 & 75 & 210 & 484 & 816 & 1173 & 1573 & 2022 \\
\cline{1-11}
\bottomrule
\end{tabular}
\end{subtable}

\begin{subtable}
\centering
        \begin{tabular}{llrrrllllll}
\toprule
 &  & Model size[mb] & Inference[$\mu$s] & Training[s] & 10 & 20 & 30 & 40 & 50 & 60 \\
Model & Max Depth &  &  &  &  &  &  &  &  &  \\
\midrule
\multirow[c]{5}{*}{FLP-XR} & 6 & 7 & 0.12 & 18 & 238 & 535 & 948 & 1265 & 1738 & 2280 \\
 & 9 & 45 & 0.26 & 33 & 193 & 488 & 860 & 1171 & 1593 & 2057 \\
 & 12 & 273 & 0.28 & 86 & 176 & 460 & 809 & \textbf{1123} & \textbf{1543} & \textbf{2003} \\
 & 15 & 1224 & 0.51 & 310 & \textbf{167} & \textbf{452} & \textbf{799} & 1133 & 1566 & 2044 \\
 & 18 & 4141 & 1.11 & 1021 & 170 & 459 & 819 & 1167 & 1610 & 2107 \\
\cline{1-11}
\bottomrule
\end{tabular}
\end{subtable}

\begin{subtable}
\centering
        \begin{tabular}{llrrrllllll}
\toprule
 &  & Model size[mb] & Inference[$\mu$s] & Training[s] & 10 & 20 & 30 & 40 & 50 & 60 \\
Model & LR &  &  &  &  &  &  &  &  &  \\
\midrule
\multirow[c]{5}{*}{FLP-XR} & 0.001 & 302 & 0.29 & 96 & 1423 & 3126 & 4888 & 6209 & 7955 & 9700 \\
 & 0.005 & 298 & 0.31 & 93 & 196 & 531 & 956 & 1233 & 1705 & 2207 \\
 & 0.01 & 273 & 0.28 & 86 & \textbf{176} & \textbf{460} & 809 & \textbf{1123} & \textbf{1543} & \textbf{2003} \\
 & 0.05 & 219 & 0.29 & 73 & 198 & 463 & \textbf{802} & 1127 & 1552 & 2025 \\
 & 0.1 & 218 & 0.26 & 71 & 211 & 478 & 823 & 1158 & 1597 & 2081 \\
\cline{1-11}
\bottomrule
\end{tabular}
\end{subtable}

\begin{subtable}
\centering
        \begin{tabular}{llrrrllllll}
\toprule
 &  & Model size[mb] & Inference[$\mu$s] & Training[s] & 10 & 20 & 30 & 40 & 50 & 60 \\
Model & Boosters &  &  &  &  &  &  &  &  &  \\
\midrule
\multirow[c]{5}{*}{FLP-XR} & 500 & 192 & 0.20 & 61 & 177 & 478 & 852 & 1152 & 1586 & 2056 \\
 & 625 & 232 & 0.23 & 73 & 175 & 466 & 821 & 1133 & 1554 & 2017 \\
 & 750 & 273 & 0.28 & 86 & \textbf{176} & \textbf{460} & 809 & 1123 & 1543 & 2003 \\
 & 875 & 310 & 0.34 & 108 & 180 & \textbf{460} & 807 & 1122 & 1540 & 1999 \\
 & 1000 & 341 & 0.35 & 111 & 183 & \textbf{460} & \textbf{806} & \textbf{1121} & \textbf{1539} & \textbf{1998} \\
\cline{1-11}
\bottomrule
\end{tabular}
\end{subtable}
\label{tab:grid_search_aegean}
\end{table*}

\end{document}